Laboratoire de Génie Industriel
**UNIVERSITY OF LA RÉUNION**

2004

# Submission of manuscript to Energy and Buildings

# A Genetic algorithm applied to the validation of building thermal models

*Philippe LAURET, Harry BOYER, Carine RIVIERE and Alain BASTIDE*

Contents:

Corresponding author:
**Philippe LAURET**
Université de La Réunion
Laboratoire de Génie Industriel, Equipe Génie Civil et Thermique de l'habitat
15 avenue René Cassin, BP 7151, 97705 Saint-Denis Cedex, Ile de la Réunion, FRANCE

tél : 02 62 93 81 27
fax : 02 62 93 86 65
email : lauret@univ-reunion.fr

# A Genetic algorithm applied to the validation of building thermal models

P. LAURET[1], H. BOYER, C. RIVIERE, A. BASTIDE

*Université de La Réunion, Laboratoire de Génie Industriel, Equipe Génie Civil et Thermique de l'Habitat*

*15 avenue René Cassin, BP 7151, 97705 Saint-Denis Cedex, Ile de la Réunion, FRANCE*

---

**Abstract**

This paper presents the coupling of a building thermal simulation code with genetic algorithms (GAs). GAs are randomized search algorithms that are based on the mechanisms of natural selection and genetics. We show that this coupling allows the location of defective sub-models of a building thermal model i.e. parts of model that are responsible for the disagreements between measurements and model predictions. The method first of all is checked and validated on the basis of a numerical model of a building taken as reference. It is then applied to a real building case. The results show that the method could constitute an efficient tool when checking the model validity.

---

## 1. Introduction

In the field of empirical validation of building thermal models, powerful techniques ranging from spectral analysis to sensitivity analysis have been developed and applied succesfully [1-7]. However, some of these methods are complex to implement, are computationnaly demanding and their theoretical basis limit their sphere of applicability [3].

[1] Corresponding author.
E-mail address : lauret@univ-reunion.fr (P. Lauret)

We have developed a simple method named ‘forcing’ [8,9] that consists in authorizing our dynamic simulation program CODYRUN [10] to integrate real (measured) hourly temperatures profiles during a simulation. It is therefore possible to impose for some given temperature nodes of the discrete building model a series of values, one for each time-step of simulation, which can be found from experiments.

These temperatures imposed at each simulation step act as Dirichlet type boundary conditions and are thus not calculated. These informations are propagated into the code according to the couplings between the different nodes of discretization.

We are therefore able to bring the boundary conditions close to the elementary models to be tested in order to find any possible error in the modelling. This simple method has proved successful in two previous studies [9,11].

Thus, compared to a non-valid global model, a first level of use is to check one by one possibly defective sub-models, by injecting into the global model measurements associated with the sub-model output and by re-analyzing the residuals.

However, this sequential step is tiresome and its cost in time can be reduced only in case of expert knowledge of the aspects related to the modelling (fine knowledge of the assumptions or sub-models couplings) as well as the ones related to the experiments (climatic sequence, location and type of sensors,...). From its sequential character, it can not lead to the optimal solution as the interactions between the different sub-models are not properly taken into account.

So, in order to overcome this stage of sequential identification, we used genetic algorithms (GAs) [12]. The latter are known to be interesting tools for stochastic optimization and also allow a global approach. GAs have found widespread application in business, scientific and engineering circles. In the field of enginneering science, GAs are mostly used to find the optimal value of physical parameters [13]. Nonetheless, this paper proposes an original contribution of GAs to the validation of building simulation codes as a GA is used to test the accuracy of individual algorithms of a global thermal model. Unlike the sequential approach, we feel that GAs can take into account the physical interactions between the different sub-models of a global building thermal model.

An application is proposed around an experimental test cell. The method enabled us to identify without ambiguity the components whose modelling was incorrect.

## 2. Genetic algorithms

*2.1. Overview of genetic algorithms*

Genetic algorithms (GAs) are search and optimization techniques based on the mechanisms of natural selection and natural genetics [12].

The basic idea of the evolution theory of Darwin is that, under the influence of external constraints, the living organisms auto-modified gradually to produce generations better and better fitted to their environment (survival of the fittest). In the same way as theory of evolution considers a population and not an isolated individual, GAs act on a population of individuals. The latter represents a parameter set to be optimized. The parameters (i.e. individuals) are encoded in a binary string (chromosome in the GA terminology). A GA modifies and updates the individuals iteratively, searching for good solutions of the optimization problem. Each iteration step is called a generation.

A GA evaluates the individuals in the population by using an objective function or fitness function in the GA terminology. The fitness function measures the performance of the individual with respect to the particular search problem. Inspired by the survival of the fittest idea, a GA maximizes the fitness value (unlike the classical techniques that minimize the objective function).

The goal of the problem is to find the individual that exhibits the best adaptation (i.e. the maximum possible value of the fitness function).

*2..2. A GA generation*

GAs evaluate the individuals in the population and then generate new (improved) solutions for the next generation. Each generation of parents will produce generation of children with an average performance better than the parent generation. Fig. 1 depicts a schematic representation of a GA generation.

Fig.1. A GA generation

The population initially chosen at random will evolve thanks to three operators : reproduction, crossover and permutation In other words, these three genetic operators take the initial population and generate successive populations that improve over time.

Reproduction consists in selecting the individuals for a new generation according to their fitness function values. This means that individuals with a higher fitness value have a higher probability to be part of the next population. The reproduction operation may be implemented through the design of a simple roulette wheel where each individual in the population has a slot weighted in proportion to its fitness value (see fig. 1).

The Crossover operation consists in exchanging genetic material (i.e. binary substrings) of two parents in order to produce fitter offspring. Crossover of two selected individuals (parents) is performed if a random number exceeds a probabilty known as probability of crossover *Pc* (to be set by the user). Otherwise, the parents are reproduced as identical.

Finally, mutation consists in modifying at random (but infrequently) the value of one or more bits of the children resulting from the crossover operation. Again, the change is made if a random number is higher than a probability known as probabilty of mutation *Pm*. The mutation operator enables the introduction of new genetic material by random to better explore the search space.

The generation process is repeated until the generation counter exceeds the maximum number of generations or a convergence criterion is met. Our specific GA procedure is represented in fig. 6. For a more detailed discussion, the interested reader should refer to [12].

The difficulties related to GAs concern mainly the setting of some control parameters (number of individuals in a population, maximum number of generations) and particularly the probabilities of crossover and mutation.

Nevertheless, GAs have proved to be robust optimization techniques in many fields [12]. Concerning the energy and building performance simulation field or the thermal engineering field, GAs have been used for optimization of the thermal behaviour of buildings [14], for the design of buildings in a multi-criterion optimization approach [15], for automatic tuning of controllers in HVAC systems [16] , for determining heat transfer coefficients of wall surface or for optimization of large solar hot water systems [17].

Summarizing the above statements, one may see that GAs operate on a population of possible solutions and search in a blind way for the best individual through the use of stochastic operators. So, regarding these specific features, we feel that GAs may proved to be particularly efficient in the search of the defective sub-models of a building thermal model.

## 3. The thermal model of the test cell and the experimental setup

### *3.1. Description of the test cell*

The survey concerns a real test cell that was erected for experimental validation of building thermal airflow simulation software [18].

The test cell is a cubic building with a single window on the south wall and a wooden door to the north. All vertical walls are identical and are composed of cement fibre and polyurethane. The roof is made of steel, polyurethane and cement fiber. The floor, made of respectively of concrete slabs, polystyrene and weight concrete, is thermically uncoupled from the ground thanks to the polystyrene. Fig. 2 shows a picture of the test cell. The case discussed in this paper is a passive one, and the split system visible in the picture is switched off.

Fig. 2. Picture of the test cell seen from North-West

### *3.2. Nodal model*

Our study relies on CODYRUN [10], a detailed multizone building airflow and thermal simulation software regrouping design and research aspects. The thermal model of CODYRUN is based on the nodal analysis [19]. Nodal analysis is a powerful method of investigation in the thermal analysis of systems. It has been used in several branches such as solar energy systems, micro-electronics or also the spatial field.

Considering the thermal behaviour of a building, its thermal state is determined by the continuous field of temperatures, concerning all points included in the physical limits of the building. The constitution of a reduced model (shown in fig. 3), with a finite number of temperatures or nodes, is possible by assuming some simplifications.

Fig. 3. Thermal discretisation of a building

Indeed, a typical model for heat transfer in a building based on the assumptions of mono-dimensional heat conduction through walls and linearized transfer equations at wall surfaces will yield a set of ordinary differential equations. To close the problem, we add the thermo-convective balance equation of dry-bulb air node and the radiative balance equation of the inside mean radiant temperature node.

More precisely and to simplify our discussion, we'll suppose that the heat conduction is treated with the help of a model constituted of a thermal resistance and 2 capacitors, the "*R2C*" model [20]. Therefore, there is no internal node inside the wall. The mathematical traduction of the thermal model of one building zone is consequently a linear system, including a certain number of equations of type (1) and (2), and one equation of type (3) and another one of type (4).

$$C_{si}\ \frac{dT_{si}}{dt} = h_{ci}\ (T_{ai} - T_{si}) + h_{ri}\ (T_{rm} - T_{si}) + K\,(T_{se} - T_{si}) + \varphi_{swi} \qquad (1)$$

$$\mathrm{C}_{\mathrm{se}}\ \frac{dT_{se}}{dt} = h_{ce}\ (T_{ae} - T_{se}) + h_{re}\ (T_{sky} - T_{se}) + K\,(T_{si} - T_{se}) + \varphi_{swe} \qquad (2)$$

$$C_{ai}\ \frac{dT_{ai}}{dt} = \sum_{j=1}^{Nw} h_{ci}\ S_j\,(T_{ai} - T_{si\,(j)} + c\,\dot{Q}(T_{ae} - T_{ai}) \qquad (3)$$

$$0 \;=\; \sum_{j=1}^{Nw} h_{ri}\ A_j\,(T_{si}(j) - T_{rm}) \qquad (4)$$

Eq. (1) and Eq. (2) traduce the respective thermal balance of the nodes of inside and outside surfaces. $N_W$ designating the number of walls of the envelope, eq. (3) is the one of the thermo-convective balance of the dry-bulb inside air temperature, taking into account airflow between inside and outside. Eq. (4) is the equation of the radiative balance of mean radiant temperature node.

This system of equations (called state equations) can be written in a state-space model of the form:

$$C.\dot{T}(t) = A.T(t) + B.U(t) \tag{5}$$

where *T(t)*, the state vector, comprises the temperatures at the nodes of the discretization mesh and *U(t)* is the vector of solicitations or input variables. *C* is a diagonal matrix of thermal capacities at the discretization nodes, *A* is a squared symetric matrix of parameters describing the thermal exchanges between nodes and *B* is a matrix of coupling parameters between the buiding and its environment. The output of the thermal model is usually the indoor dry air temperature $T_{ai}$.

In other words, each state equation of this linear system represents the thermal behavior of the considered node. It integrates the models of the various physical phenomena and hence the physical coupling with the others nodes.

The mathematical model of a building is considered a global model as it involves several so-called elementary models. Therefore, the validation procedure consists in verifying these sub-models (or individual algorithms) but also their couplings as a building model can be seen as a combination of elementary models.

Finally, the modelling of the test cell led to a grid of 23 nodes. Some of these nodes are listed in table 1. An implicit finite differences scheme was used for the numerical resolution of eq. (5). Therefore, the discretization of eq. (5) yields the following linear system:

$$M.T^{n+1} = V \iff \begin{pmatrix} m_{1,1} & \cdots & m_{1,23} \\ \vdots & & \\ m_{i,1} & \ddots & m_{i,23} \\ \vdots & & \\ m_{23,1} & \cdots & m_{23,23} \end{pmatrix} \begin{pmatrix} T_1 \\ T_2 \\ \vdots \\ T_i \\ \vdots \\ T_{22} \\ T_{23} \end{pmatrix}^{n+1} = \begin{pmatrix} v_1 \\ \vdots \\ . \\ v_i \\ . \\ v_{22} \\ v_{23} \end{pmatrix} \tag{6}$$

where *n* is the simulation time step.

Considering this numerical resolution, heat flux (taken at a specified depth) under the floor is deemed null. This assumption appears to be reasonable in the present case since the floor is thermally uncoupled from the ground (through the use of an insulation layer).

### *3.3. Experimental setup*

Experiments were carried out in march 2000. A weather station near the test cell provided the necessary solicitations to our building model. Data were sampled every minute and averaged every 0.25 hour. Fig. 4 shows the instrumentation of the test cell.

Fig. 4. Instrumentation of the test cell

In addition to the traditional measurements (surface temperature, indoor dry-bulb temperature, humidity), a specific attention has been put in the instrumentation of the ground. Consequently, temperature sensors have been inserted in the weight concrete supporting the cell at respectively 10 cm and 5 cm of depth.

Table 1 lists the nodes for which a measurement is available.

Table 1: List of the available measurements

## 4. The forcing method

The idea of combining a mathematical model with physical measurements has sound theoretical basis, particularly in the field of modern control theory for the design of state-space observers [21] or the so-called software sensors. Successful applications of the concept can be found in [22,23].

Regarding our study, at each time step of the simulation program, instead of resolving the state equation, the temperature of the node for which a measurement is available is set to this measured value. More precisely, the set of linear equations given by eq. (6) is rewritten under the following form (where, for instance, the $i^{th}$ temperature node $T_i$ is set to its counterpart measured value $T_{mes}$):

$$M^*.T^{n+1} = V^* \Leftrightarrow \begin{pmatrix} m_{1,1} & \cdots & \cdots & m_{1,23} \\ \vdots & & & \\ 0 & \ldots & 1 & \ldots & 0 \\ \vdots & & & \\ m_{23,1} & \cdots & \cdots & m_{23,23} \end{pmatrix} \begin{pmatrix} T_1 \\ T_2 \\ \vdots \\ T_i \\ \vdots \\ T_{22} \\ T_{23} \end{pmatrix}^{n+1} = \begin{pmatrix} v_1 \\ \vdots \\ . \\ T_{mes} \\ . \\ v_{22} \\ v_{23} \end{pmatrix} \quad (7)$$

In other words, a Dirichlet condition is imposed on the considered node. This simple method has been implemented in our building thermal simulation software CODYRUN.

Indeed, in addition to the building description file and meteorological input file, a third CODYRUN input file (file extension .mes) comprises the time series of temperature of the nodes for which a measurement is available.

A window enables the CODYRUN user to select the nodes in order to check one by one sub-models (by injecting into the global model measurements associated with the sub-model and by re-analyzing the residuals). As mentioned above, this method has proved successful in two previous studies [9,11].

However, this sequential method is far from being optimal  and one has to automate the search for defective sub-models.

## 5. GA applied to the validation of building thermal models

As stated above, the previous sequential method that consists in forcing one by one the nodes of the discrete model cannot lead to the optimal solution. Interactions between the nodes due to the physical couplings must be considered in order to correctly treat the problem. This fact is reinforced when dealing with complex buildings (i.e multi-zone buildings).

We feel that GA could be the solution for finding the nodes and hence the corresponding sub-models that are responsible for the the disagreements between measurements and model predictions.

Since the test cell is not a complex one and since the amount of available measurements is limited, the purpose of the present application is to demonstrate the efficiency of the GA approach. However, the extension to a full-sized building will be straitghforward.

In the present application, an individual is a binary string of length 22 which indicates the nodes of the discrete model to impose or "to force". Node 23 corresponding to the indoor dry air temperature does not belong to the chromosome as the optimization concerns this variable. If the *i*th bit of the binary string is set to one, it indicates that the corresponding node is to be forced. As an example, fig.5 indicates that six node temperatures i.e. 1, 5, 10, 16 and 20 are imposed during a simulation.

Fig. 5. Example of individual or chromosome : nodes 1,5,10,16, 20 are imposed.

We recall that in our experimental configuration, the nodes for which we had a measurement are nodes T1, T7, T9, T16 and T17 (see table 1). Measurements of these two last nodes were made possible thanks to the thermocouples inserted in the weight concrete.

Five measurements are sufficient to illustrate the efficiency of the method on a simple building as the test cell. Again, it will be of course desirable to apply this method on a complex building with as much measurements available as possible.

The goal of the optimization problem is to find the combination of temperature node measurements that minimizes the following objective function i.e. the sum squares of the residuals:

$$J = \sum_{i=1}^{N} \left( T_{ai}^{\exp,i} - T_{ai}^{simul,i} \right)^2 \qquad (7)$$

where $N$ represents the number of experimental points, $T_{ai}^{\exp}$ is the measurement of the indoor dry air temperature and $T_{ai}^{simul}$ the model response computed by CODYRUN. In our GA application, it is important to note that the model response $T_{ai}^{simul}$ is determined by solving the mathematical model (Eq. 5) but with possibly some node temperatures of the model imposed by the GA. Fig. 6 illustrates the interaction between CODYRUN and the GA.

Fig. 6. Diagram flow chart of the GA procedure

Notice that in the GA terminology $J$ is called the objective score or simply score. Notice also that such a definition of $J$ allows to exhibit the state variables (i.e. node temperatures) that influence the model response ($T_{ai}^{simul}$).

As the GA operators are designed to maximize the fitness function $f$, the above minimization problem has to be transformed into a maximization one. This can be done by using the following transformation:

$$f = \frac{1}{1+J} \tag{8}$$

The individuals having the best adaptation (in other words the best fitness $f$) will be those for whom $J$ will be minimum. If, by setting the node temperature of the discrete model to its measured value, we observe a significant reduction in the score $J$ then we can reasonably guess that the sub-model attached to this node is a source of error. In other words, the existence of a strong disparity (on the indoor dry air temperature) between the models integrating for one the series of computed values and for the other the series of measurements enables us to detect a problem of modelling of the phenomena related to this node. Nonetheless, the power of the method relies on high quality measurements.

As mentioned above, GAs are robust stochastic search methods. One may say that the GA stretchs the building thermal model. As a consequence, they allow to test the accuracy of individuals algorithm  and to exhibit the defective ones.

D

## 6. Results

### *6.1. Verification of the algorithm with regard to a reference model*

A verification of the algorithm was carried out by taking as experimental pseudo-data the simulation results of the computer program on a basic case (or reference model) of the test cell.  Starting from this reference model, we have tested three cases by  disturbing the sub-models listed in table 2. Table 3 shows the results of the GA search procedure.

Table 2: Three testing cases

Table 3: The corresponding results

For the first case, the algorithm indicates that if one replaces the equation relative to node 7  by the associated measurement, it turns out that the performance of the model is better. In other words, the AG points directly towards the defective sub-model. The same comment can be made for the third case. However, case 2 leads to a particular comment. Indeed, the algorithm converges towards an individual where none of the nodes is replaced by its measurement. In that case, one can see that the resulting score is very important. This appears completely logical from the role of coupling played by the coefficient *hci* between indoor surface temperatures and the indoor dry air temperatures $T_{ai}$. With respect to the disturbance under consideration in this case (*hci* from 5 to 0.1), the link between the variable ($T_{ai}$) on which is made the optimization and the others nodes appears to be strongly reduced (see eq. 3). Therefore, this leads the GA to logically not identify measurements that improve the objective function.

Apart from this last case, the convergence of the algorithm towards the defective sub-models allows the verification of the procedure.

Further, this verification step has helped to validate the choice of the parameters of the genetic algorithm. We fixed the size of the population at 30 individuals and took for the probabilities of crossover and mutation the following values $Pc$=0.8 and $Pm$=0.03. These values are close to the traditional choices carried out in the literature [12].

### *6.2. Application of the method on the test cell*

The following stage was to test globally our algorithm on the test cell. A five day-long experiment carried out in March 2000 enabled us to test the GA procedure search for defective sub-models. At the end of 400 generations or iterations, the algorithm converged towards the following best individual:

Fig. 7. Best individual

The procedure thus seems to indicate that the defective sub-models are related to nodes 1, 9 and 17. Notice that the GA did not propose the node 16. The relative performance of each node is listed in table 4.

Table 4: Scores of the best individual

A previous parametric sensitivity analysis [9] pointed out the relative importance of parameters related to the door. As the door is the least insulated element of the test cell, a particular attention has to put on the modelization of this component. The same study has also put forward the influence of roof-effects (through the solar absorption coefficient) on the global thermal behavior of the cell. The best performance is obtained when forcing node 17. Indeed, for the present case, the assumption of null heat flux when dealing with the conduction process through the floor appears to be too strong.

Focusing on this particular node, two previous studies [9,11] based on a parametric sensitivity analysis highligthed the importance of ground related effects and  emphasized the need to properly identify the physical phenomena of heat conduction through the floor and consequently the sub-model attached to it.

Comparison between the temperature measured at 10 cm in weight concrete under the cell with that predicted by the model put forward an identical tendency but with a shift ranging between 0.5 and 1°C (see fig. 8). Forcing this node leads to an improvement of the model and specifically an improvement of the mean of the residuals (i.e. steady state response of the model ) (see table 5)

Fig. 8.  Temperatures measured in the ground at different depth

Table 5: Statistical characteristics of the residuals

As one can see, the proposed method constitutes a first step by locating defective sub-models or faulty individual algorithms. A second step will be to identify parameters of the sub-models that are at the origin of the error.

## 7. Conclusion

The purpose of the paper has been to illustrate the feasibility of the method on a simple building. The power of GAs (search for an optimal solution starting from a population of possible solutions, stochastic exploration of the search space,…) would be more effective on a complex building. Thus, the next step will be to extend the work from test cells to full-sized buildings.

The proposed method constitutes a first step by locating defective sub-models of a global building model. Therefore, on can say that the procedure allows a spatial diagnosis of the building model and we feel that the procedure could be a valuable tool for the analyst or the modeller to search in the good direction in order to improve his model. So, the goal of the proposed approach is to analyze the discrete nodal model and to pinpoint the defective sub-models.

A second step (not provided by the tool) will be to find the parameters of the sub-models that are responsible for the disagreements between simulations and measurements. For that step, a parametric sensitivity analysis would be of great help.

***NOMENCLATURE***

**A** square matrix

**B** vector

**C** diagonal matrix

$C$ thermal capacity

$c$ specific heat

$h$ heat transfer coefficient

$K$ thermal conductivity

$\varphi$ radiation flux density

$\dot{Q}$ mass flow rate

$S$ surface area

$T$ temperature

**U** matrix

Subscripts :

| | |
|---|---|
| *ae* | air ouside |
| *ai* | air inside |
| *se* | surface outside |
| *si* | surface inside |
| *ce* | outside convection |
| *ci* | inside convection |
| *lwe* | long wave outside |
| *re* | outside radiation |
| *ri* | inside radiation |
| *rm* | inside radiant mean |
| *swe* | short wave outside |
| *swi* | short wave inside |

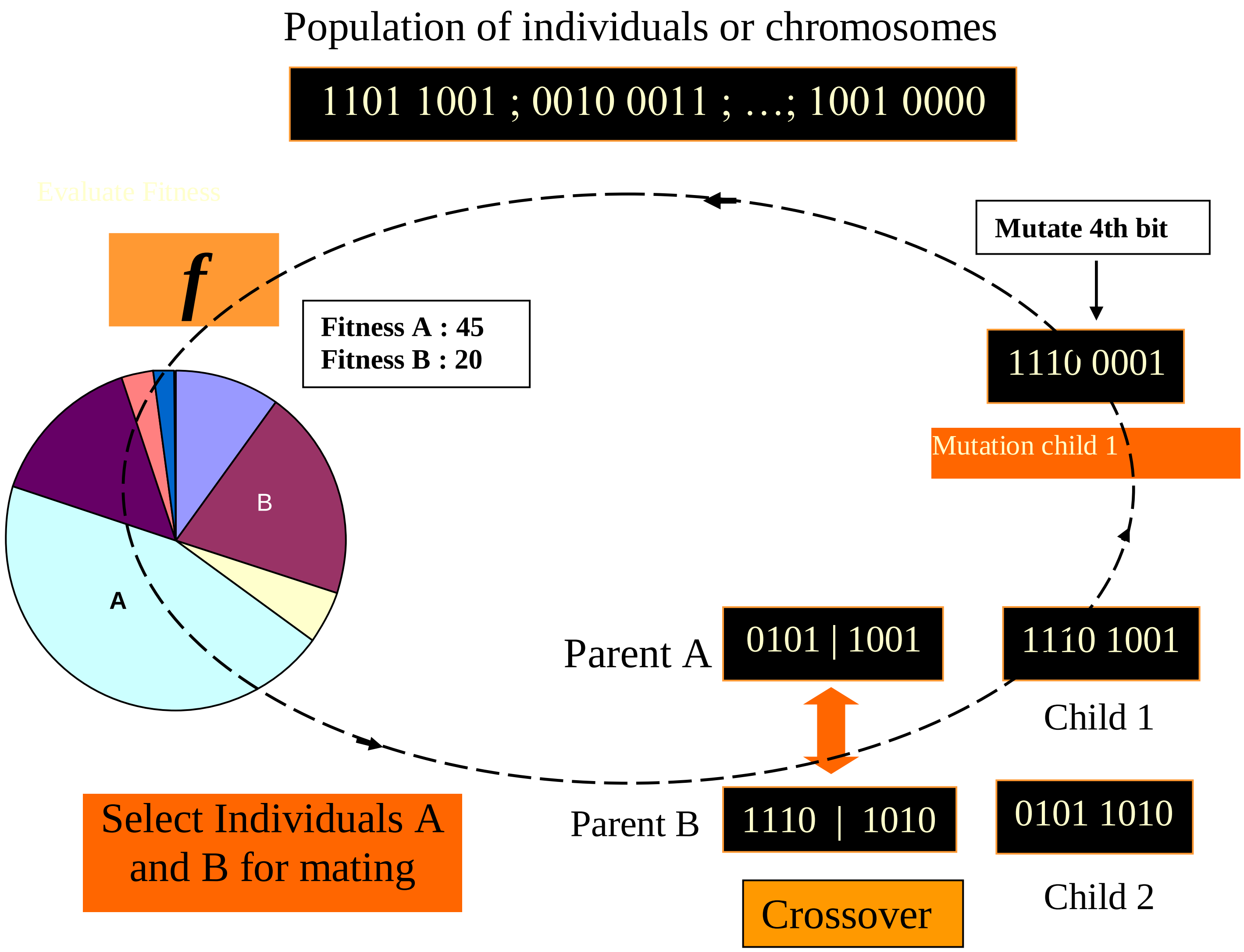

Fig.1. A GA generation

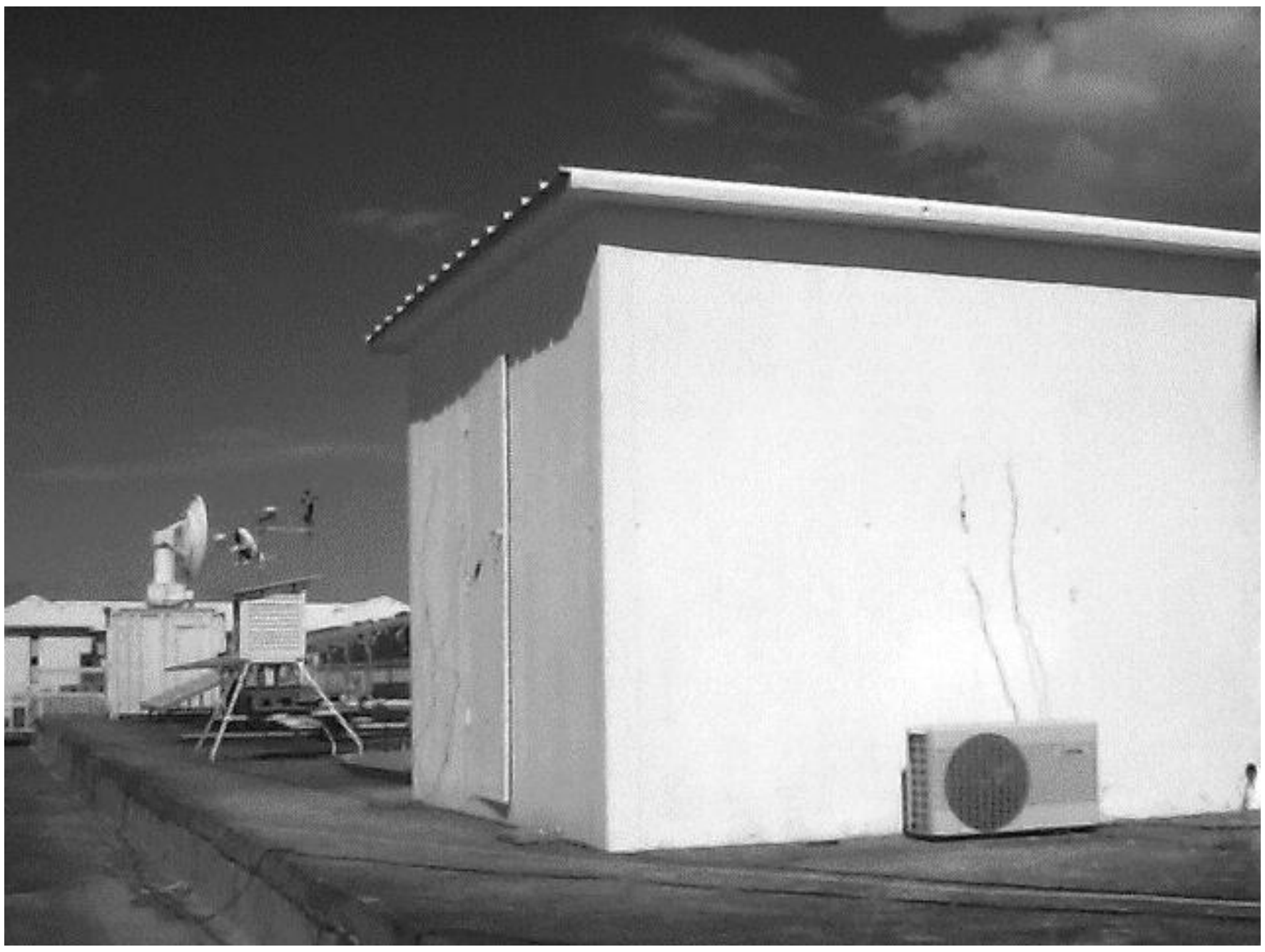

Fig. 2. Picture of the test cell seen from North-West

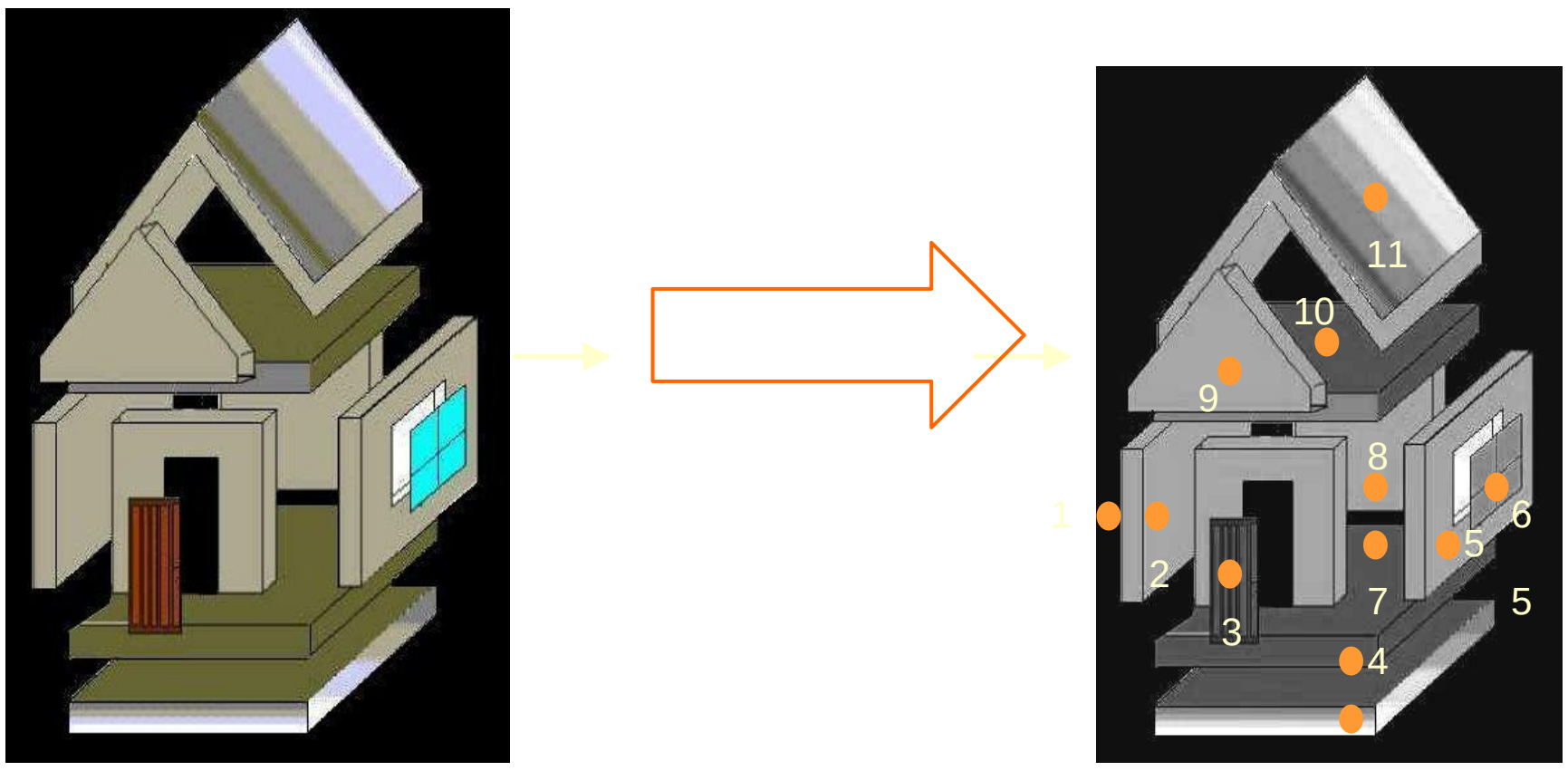

Fig. 3 : Thermal discretisation of a building

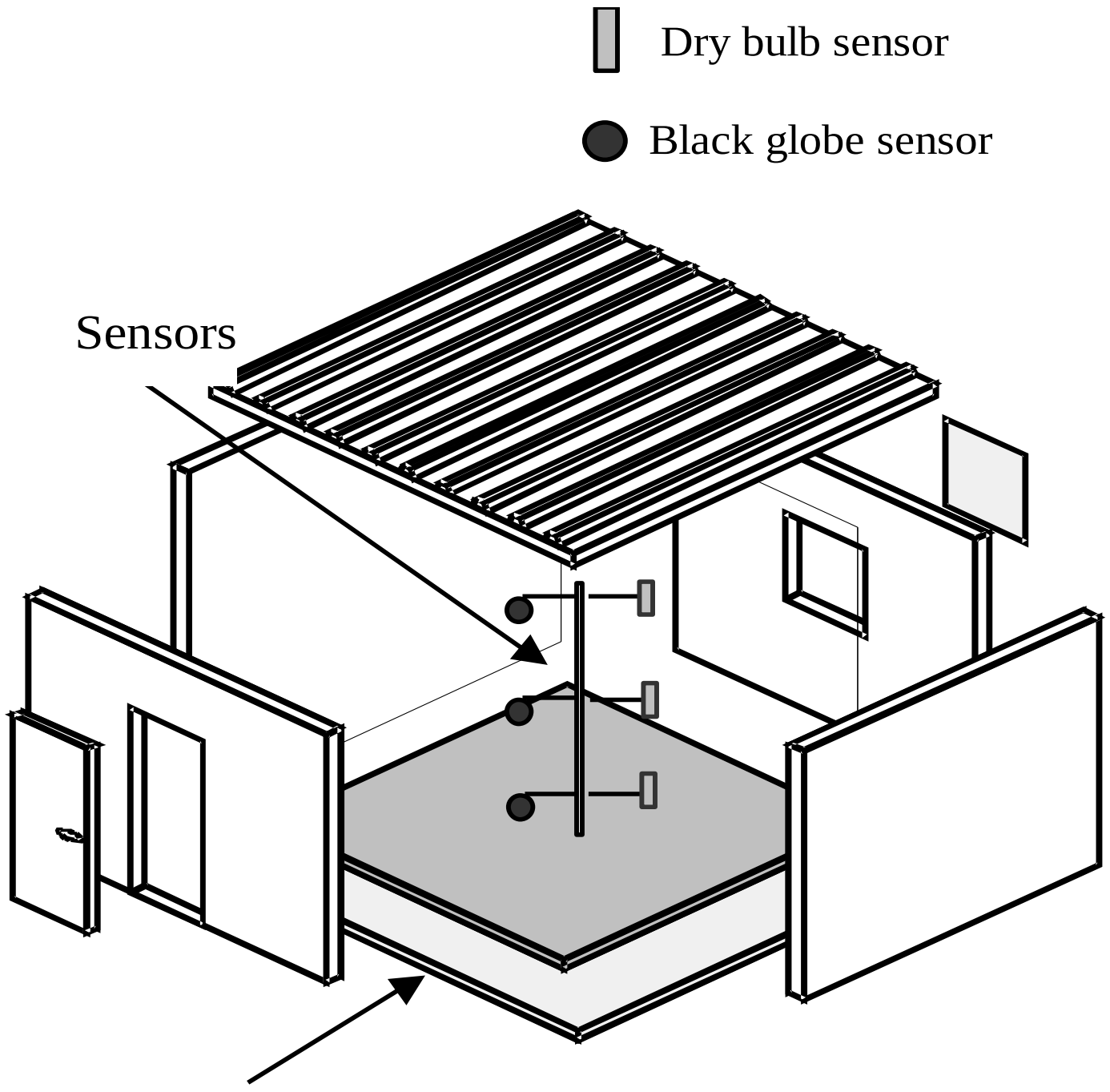

Fig. 4. Instrumentation of the test cell.

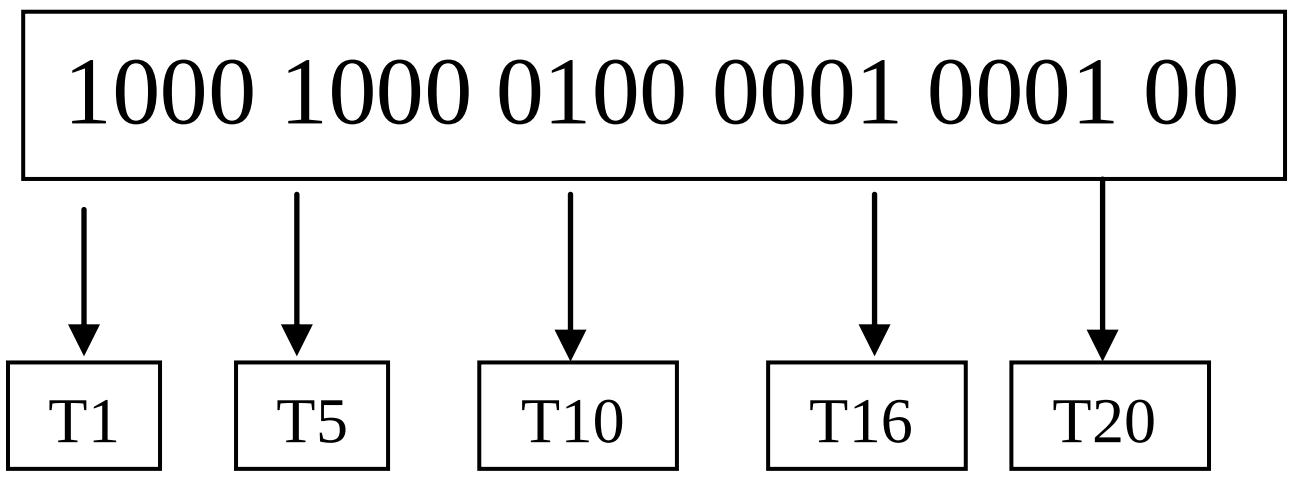

Fig. 5. Example of individual or chromosome: nodes 1,5,10,16, 20 are imposed.

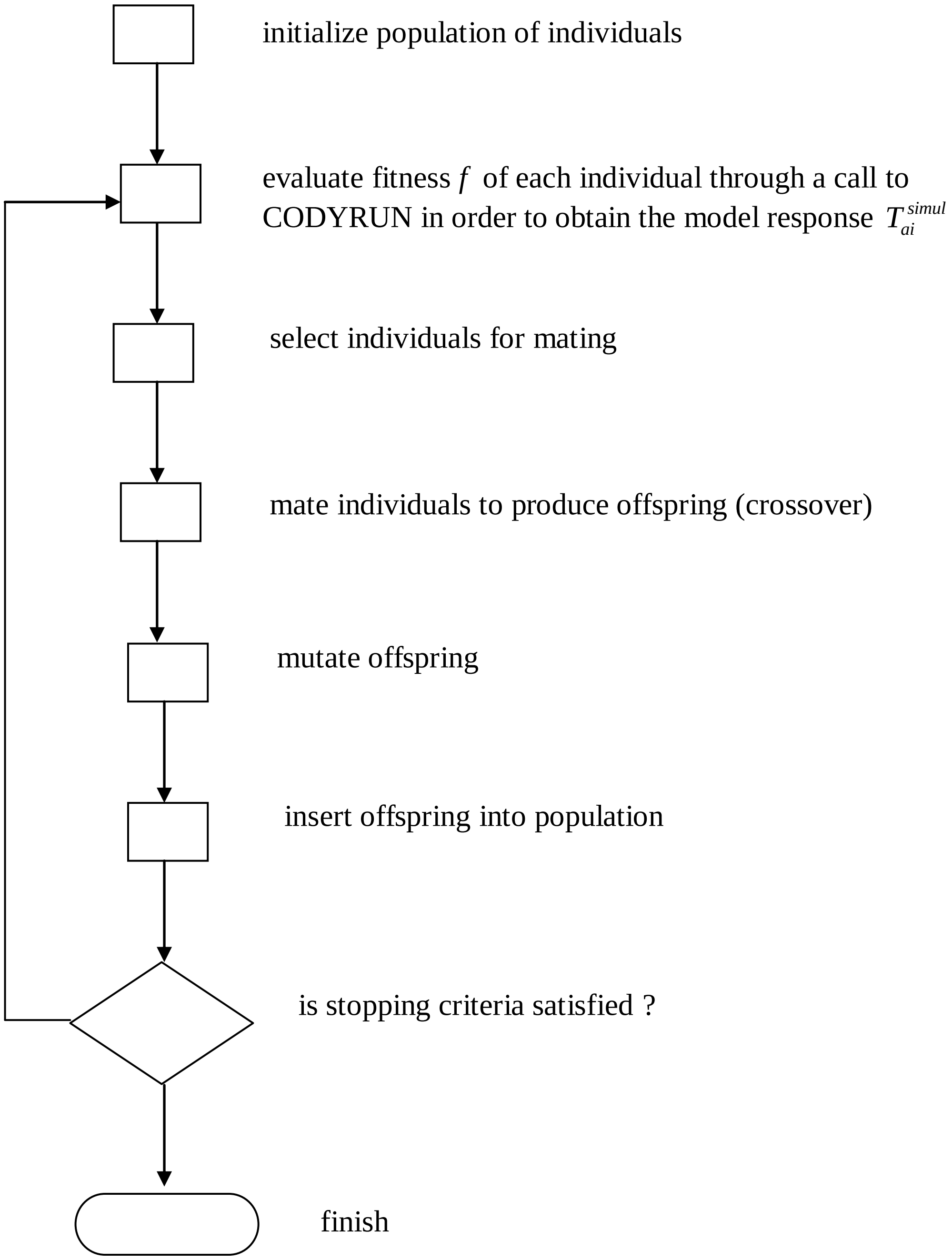

Fig. 6. Diagram flow chart of the GA procedure

| |
|---|
| 1000 0000 1000 0000 1000 00 |

Fig. 7. Best individual

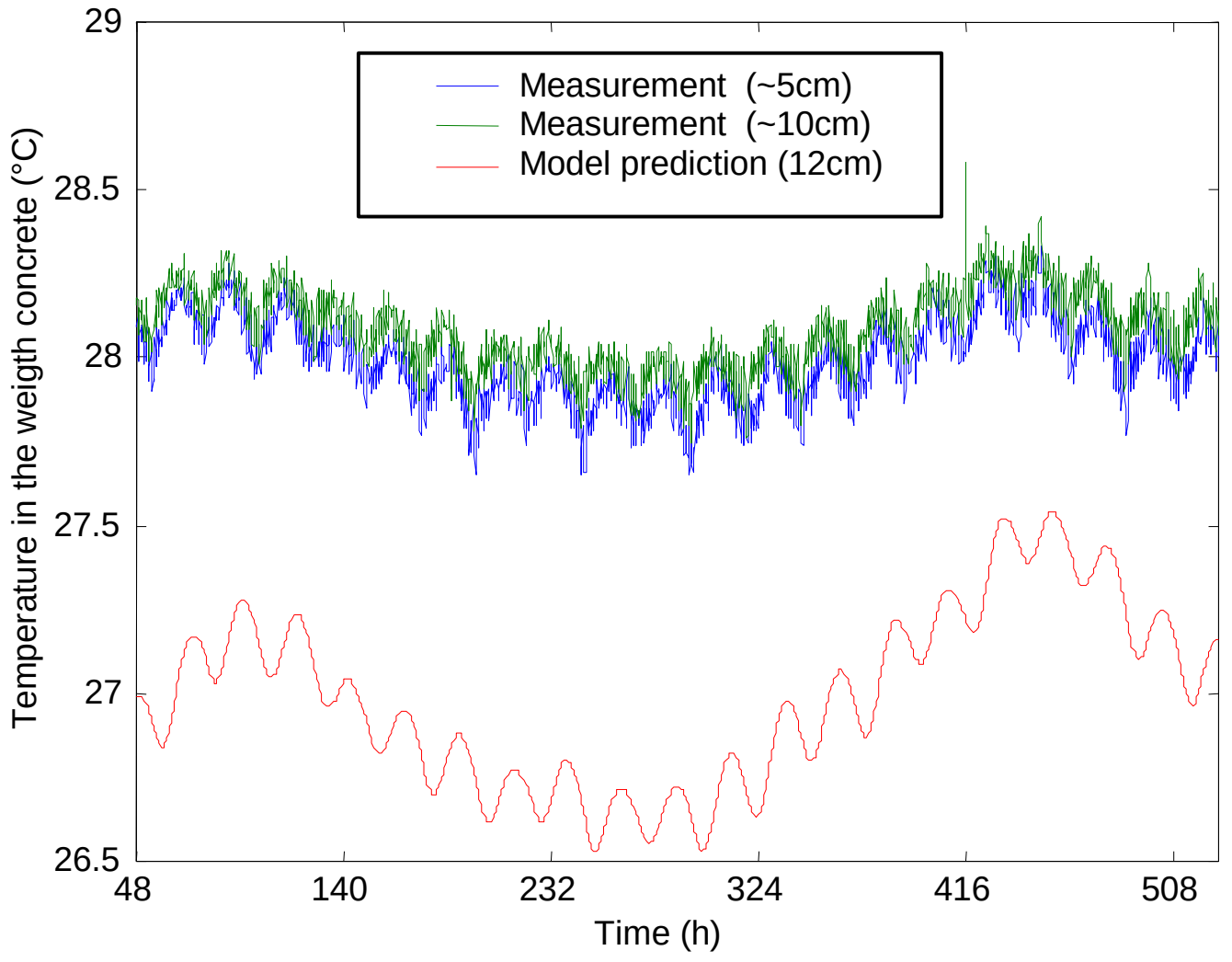

Fig. 8. Temperatures measured in the ground at different depth

Table 1

List of the available measurements

| Node Number | Type and location of the node |
|---|---|
| 1 | Indoor surface node of east wall |
| 7 | Indoor surface node of the door |
| 9 | Indoor surface node of the roof |
| 16 | Internal node of ground wall (5 cm) |
| 17 | Internal node of ground wall (10 cm) |
| 23 | indoor dry air temperature |

Table 2

Three testing cases

| Case# | sub-model | modification (base case to new value) |
| --- | --- | --- |
| 1 | Conductive model of the door | $\lambda$ thermal conductivity (0.23 to 0.78) |
| 2 | Indoor convective model | *hci* indoor convective heat exchange coefficient (5 to 0.1) |
| 3 | Radiative model of the roof | $\alpha$ absorptivity of the roof (0.3 to 0.9) |

Table 3

The corresponding results

| Case# | Nodes exhibited by the best individual | Score $J(°C^2)$ when the node is forced | Score *J* when the node temperature is *not* replaced by the measurement (i.e. without forcing) |
|---|---|---|---|
| 1 | Node 7 | 14.06 | 177.15 |
| 2 | No Nodes forced | 197.66 | 197.66 |
| 3 | Node 9 | 7.04 | 577.05 |

Table 4

Scores of the best individual

| Node | Score $J$ ($°C^2$) |
|---|---|
| 1 | 50.60 |
| 9 | 27.80 |
| 17 | 13.08 |
| Without forcing | 122.50 |

Table 5

Statistical characteristics of the residuals

| Node | Before forcing | After forcing |
|---|---|---|
| Mean | 0.232 °C | -0.012 °C |
| S.D. | 0.377 °C | 0.340 °C |

.